\documentclass[conference]{IEEEtran}
\usepackage{cite}
\usepackage{amsmath,amssymb,amsfonts}
\usepackage{algorithmic}
\usepackage{graphicx}
\usepackage{textcomp}
\usepackage{xcolor}
\usepackage{algorithm}
\usepackage{algorithmic}
\usepackage{amsmath}

\usepackage{tikz}
\usetikzlibrary{positioning, arrows.meta, calc, fit, backgrounds}
\usepackage{amsthm}
\theoremstyle{remark}

\def\BibTeX{{\rm B\kern-.05em{\sc i\kern-.025em b}\kern-.08em
T\kern-.1667em\lower.7ex\hbox{E}\kern-.125emX}}

\usepackage{fancyhdr}
\begin{document}

\title{Adaptive Quantized Planetary Crater Detection System for Autonomous Space Exploration}

\author{
  \IEEEauthorblockN{Aditri Paul\IEEEauthorrefmark{1}, Archan Paul\IEEEauthorrefmark{2}}
  \IEEEauthorblockA{\IEEEauthorrefmark{1}\textit{Manipal University Jaipur}, India. Email: aditripaul@gmail.com}
  \IEEEauthorblockA{\IEEEauthorrefmark{2}\textit{Computational Scientist}, \textit{India}. Email: archanpaul@gmail.com}
}

\maketitle
\thispagestyle{fancy}

\begin{abstract}
  Autonomous planetary exploration demands real-time, high-fidelity environmental perception. Standard deep learning models require massive computational resources. Conversely, space-qualified onboard computers operate under strict power, thermal, and memory limits. This disparity creates a severe engineering bottleneck, preventing the deployment of highly capable perception architectures on extraterrestrial exploration platforms. In this foundational concept paper, we propose the theoretical architecture for the Adaptive Quantized Planetary Crater Detection System (AQ-PCDSys) to resolve this bottleneck. We present a mathematical blueprint integrating an INT8 Quantized Neural Network (QNN) designed specifically for Quantization Aware Training (QAT). To address sensor fragility, we mathematically formalize an Adaptive Multi-Sensor Fusion (AMF) module. By deriving the exact integer requantization multiplier required for spatial attention gating, this module actively selects and fuses Optical Imagery (OI) and Digital Elevation Models (DEMs) at the feature level, ensuring reliable perception during extreme cross-illuminations and optical hardware dropouts. Furthermore, the architecture introduces anchor-free, center-to-edge regression heads, protected by a localized FP16 coordinate conversion, to accurately frame asymmetrical lunar craters without catastrophic integer truncation. Rather than presenting physical hardware telemetry, this manuscript establishes the theoretical bounds, structural logic, and mathematical justifications for the architecture. We outline a rigorous Hardware-in-the-Loop (HITL) evaluation protocol to define the exact testing criteria required for future empirical validation, paving the way for next-generation space-mission software design.
\end{abstract}

\begin{IEEEkeywords}
  Crater Detection, Deep Learning, Model Quantization, Attention Mechanisms, Computer Vision, Artificial Intelligence (AI), Real-Time Object Detection, Sensor Fusion, Edge AI.
\end{IEEEkeywords}

\section{\textbf{Introduction}}

  The reliability for autonomous planetary navigation fundamentally depends on how effectively a platform can interpret its environment in real time through its specialized onboard computers and space-grade sensors. From the early Apollo landings to the Artemis missions, immediate environmental perception remains a prerequisite for mission success. Planetary craters function as fundamental geological records and act as highly reliable fiducial markers for localization, terrain mapping, and hazard avoidance \cite{2}. Real-time crater detection algorithms must therefore remain adaptive to the operating environment and stay computationally lean. This processing efficiency is what ultimately enables precise, autonomous trajectory tracking for landers and rovers operating without Earth-based intervention.

  Achieving this reliability in deep space introduces a severe contradiction. Modern high-accuracy deep learning models require massive computational resources. Conversely, space-qualified hardware operates under strict power, thermal, and memory constraints. Current radiation-hardened onboard computers possess highly restricted computation and memory capacities and they need to operate within a rigid power envelopes. This specific disparity between algorithmic demand and hardware capability exposes a primary operational bottleneck.

  While deep learning and sensor fusion are mature fields in terrestrial computing, their transition to space-qualified onboard hardware poses unique challenges. Most single-modality detectors falter when confronted with extreme shadows or specular glare. We therefore frame this theoretical transition as a rigorous optimization problem: extraction of high-density feature representations from high-bandwidth sensor streams without exceeding the finite throughput of space-grade onboard computers.

  Addressing this optimization problem involves leveraging theoretical insights from recent progress in resource-constrained terrestrial networking. Specialized algorithms developed for Integrated Sensing and Communication (ISAC) networks have already established that sensing performance can be maximized even under rigid power limitations \cite{23}. Furthermore, deep learning frameworks designed for multi-target radar estimation demonstrate that prediction latency can be reduced without compromising accuracy, even in environments with highly variable signal-to-noise ratio (SNR) levels \cite{22}. We treat planetary crater detection as a parallel resource-allocation task. Since identifying crater boundaries under harsh lunar lighting is essentially a low-SNR problem, our objective is to define the theoretical bounds for geometric feature extraction while strictly adhering to the finite hardware capacities of space-grade systems.

  To bridge the gap between high-fidelity hybrid vision and radiation-hardened computation hardware, we introduce the architectural framework for the Adaptive Quantized Planetary Crater Detection System (AQ-PCDSys) [10]. We present this manuscript strictly as a foundational position paper and theoretical blueprint. Every proposed design choice reflects the absolute necessity for high-performance inference on processors with limited computation power and memory. Rather than attempting to compress heavy foundation models post-training, AQ-PCDSys operates as a ground-up, hardware-aware theoretical concept. Because we scoped this work as a foundational blueprint, our immediate focus rests entirely on establishing rigorous mathematical and structural justifications. Explicit empirical benchmarking, physical ablation studies, and final hardware telemetry remain deferred to subsequent physical testing phases and future papers.

  The primary theoretical contributions of this paper are as follows:

  \begin{itemize}
    \item We propose the architectural blueprint for a dual-backbone (for optical and topographical data streams), QAT-native perception system. By utilizing depthwise factorization and mathematically folding normalization parameters prior to quantization, the design strictly bounds computation and memory complexity to optimize execution within the rigid physical envelopes of radiation-hardened onboard hardware.
    \item We mathematically formalize the Adaptive Multi-Sensor Fusion (AMF) module. The framework derives the exact integer requantization multiplier required to execute multiplicative spatial attention gating without hardware overflow, paired with a pointwise channel compression step to resolve memory bottlenecks during INT8 tensor concatenation.
    \item We define an anchor-free, center-to-edge regression architecture that anchors distance predictions directly to fused topographical features to physically accommodate asymmetrical lunar craters. This is paired with a localized FP16 dequantization step that converts integer distances into Cartesian coordinates, preventing geometric degradation during final hazard resolution.
    \item We outline a strict Hardware-in-the-Loop (HITL) experimental protocol to define the empirical criteria necessary to benchmark the system against existing baseline architectures in future physical trials.
  \end{itemize}

  \section{\textbf{Related Work}}

  The earliest efforts in planetary crater detection relied on classical image processing techniques like edge detection, morphological operations, and template matching \cite{12}, \cite{13}. These rule-based techniques provided a necessary baseline, but their reliability remains too limited for the dynamic, unpredictable planetary environment. A slight shift in solar angle or a patch of deep shadow severely degrade the accuracy of the predictions. Morphological variety among craters presents a persistent challenge. Because systems tuned for distinct, well-defined rims frequently fail to classify/detect when encountering degraded or partially buried formations and the researchers are forced into exhausting loops of manual recalibration.

  Extraction of reliable features from unpredictable environments is a persistent challenge across multiple domains. In the context of radar systems, recent convolutional neural network methods for multi-target estimation have proven highly effective at reducing prediction latency while maintaining accuracy across widely varying signal-to-noise ratio (SNR) regimes [22]. This research presents a compelling theoretical parallel to our own challenges in space exploration. Identifying crater rims under extreme cross-illumination is fundamentally a low-SNR problem. The true geometric signal is frequently buried within the noise of deep shadows or specular glare. This parallel justifies our proposed architectural shift toward modules that can actively filter environmental noise in real time.

  Recent progress in highly matured convolutional neural networks (CNNs) revolutionized feature extraction. Exceptional accuracy can be achieved using two-stage detectors like Faster R-CNN \cite{11} by isolating potential regions of interest first. However, their massive computational cost makes them unsuitable for the real-time constraints of an active automated landing sequence and planetary rover functionalities. This realization pushed the research community toward single-stage detectors like the YOLO series \cite{6}, \cite{7}. While these visual-data-only models perform well under ideal conditions, their reliability severely degrades when the optical data stream is compromised by planetary environmental factors, such as extreme albedo variations, or physical operating conditions, such as dust and vibrations during the terminal landing sequence.

  Deploying these single-stage architectures on commercial edge devices, such as the NVIDIA Jetson or Google Edge TPU, often reveals a significant drop in accuracy when applying standard post-training quantization. To address this precision loss, many current optimization frameworks favor 8-bit floating-point (FP8) precision, as it preserves representational fidelity better than fixed-point mathematics. The theoretical advantages of such approaches are well-documented. Unfortunately, standard space-qualified flight computers and radiation-hardened processors currently lack native FP8 arithmetic logic units. Consequently, the AQ-PCDSys blueprint specifically targets strict INT8 fixed-point execution to ensure immediate theoretical compatibility with available flight hardware.

  Meanwhile, the broader computer vision community has shifted toward immense foundation models that require massive computation resources. Current research has delved into Segment Anything Model (SAM) pipelines and specialized Vision-Language Models (VLMs) optimized for terrain analysis, both of which demonstrate remarkable zero-shot generalization on lunar data. Yet, as noted in comprehensive crater-detection surveys, the actual transition of these massive systems onto edge deployed platforms remain an unrealized goal for space missions. Their reliance on heavy FP32 matrix operations fundamentally violates the strict thermal and power envelopes of space-grade onboard computers. Additionally, these models overly rely on optical imagery and largely ignore the critical concept of sensor fusion.

  Finally, crater detection does not operate in isolation. It functions as the real-time perception front-end for broader localization systems. Architectures such as LunarNav \cite{20} and ShadowNav \cite{21} demonstrate the absolute operational necessity of sensor fusion for navigating extreme environments. Resilient multi-modal designs often utilize adaptive routing networks, such as mixture-of-experts, to maintain performance during sensor failures. We contextualize our own Adaptive Multi-Sensor Fusion (AMF) module and its internal Adaptive Weighting Mechanism (AWM) against these late-fusion precedents. The proposed AQ-PCDSys framework outlines a computationally lighter alternative to heavy routing networks by executing fusion at the feature level using simple, quantized spatial gating.

\section{\textbf{System Architecture and Methodology}}

  We propose the AQ-PCDSys theoretical framework to resolve specific technical gaps identified in terrestrial perception models. The system is conceptually structured as a unified architecture where each component supports the others. The proposed design conserves computational capacity through strict quantization. It then allocates that spare computational power toward an Adaptive Fusion Module which integrates the optical and terrain data-stream intelligently. This tightly coupled theoretical design ensures the necessary balance between efficiency, stability, and accuracy.

  \subsection{\textbf{System Overview and Operational Context}}

    AQ-PCDSys is designed to operate as an integrated perception module within a broader autonomous outer space navigation framework. Its primary objective is to resolve the bottleneck of executing intensive AI models on sparse, resource-constrained onboard computers.


\definecolor{optblue}{RGB}{214, 234, 248}
\definecolor{demorange}{RGB}{250, 229, 211}
\definecolor{fusepurple}{RGB}{232, 218, 239}
\definecolor{navgreen}{RGB}{212, 239, 223}
\definecolor{sysgray}{RGB}{236, 240, 241}
\definecolor{alertred}{RGB}{250, 219, 216}

\begin{figure}[htbp]
  \centering
  \resizebox{0.80\columnwidth}{!}{%
    \begin{tikzpicture}[
        node distance=0.6cm and 0.4cm,
        halfnode/.style={rectangle, draw=black!70, rounded corners, thick, minimum width=3.8cm, minimum height=1.2cm, align=center, font=\scriptsize},
        fullnode/.style={rectangle, draw=black!70, rounded corners, thick, minimum width=8.0cm, minimum height=1.0cm, align=center, font=\scriptsize},
        arrow/.style={->, >=stealth, thick, draw=black!70}
      ]

      \node[halfnode, fill=optblue] (cam) {\textbf{Optical Camera} \\ Texture \& Albedo};
      \node[halfnode, fill=demorange, right=of cam] (lidar) {\textbf{LiDAR / Stereo} \\ Elevation Data};

      \node[halfnode, fill=optblue, below=of cam] (img) {\textbf{Optical Imagery} \\ (Grayscale)};
      \node[halfnode, fill=demorange, below=of lidar] (dem) {\textbf{DEM Data} \\ (Topography)};

      \coordinate (mid_data) at ($(img)!0.5!(dem)$);
      \node[fullnode, fill=fusepurple, below=2.0cm of mid_data] (pre) {\textbf{Input Processing} \\ Resizing \& Normalization};
      \node[fullnode, fill=fusepurple, below=of pre] (qnn) {\textbf{AQ-PCDSys Core} \\ QAT Neural Network};
      \node[fullnode, fill=fusepurple, below=of qnn] (det) {\textbf{Feature Detection} \\ Bounding Boxes, Confidence, \& Class Labels};

      \node[fullnode, fill=sysgray, below=0.8cm of det] (res) {\textbf{Telemetry} \\ Real-Time Hazard Feed};

      \node[fullnode, fill=navgreen, below=1.0cm of res] (nav) {\textbf{Navigation Module} \\ Path Planning};
      \node[fullnode, fill=navgreen, below=of nav] (loc) {\textbf{Localization Module} \\ Crater-Based SLAM};
      \node[fullnode, fill=navgreen, below=of loc] (haz) {\textbf{Hazard Avoidance} \\ Safe Landing Sequence};

      \draw[arrow] (cam) -- (img);
      \draw[arrow] (lidar) -- (dem);

      \draw[arrow] (img.south) -- ([xshift=-2.1cm]pre.north);
      \draw[arrow] (dem.south) -- ([xshift=2.1cm]pre.north);

      \draw[arrow] (pre) -- (qnn);
      \draw[arrow] (qnn) -- (det);
      \draw[arrow] (det) -- (res);

      \draw[thick, draw=black!70] (res.west) -- ++(-0.6,0) coordinate (bus);
      \draw[arrow] (bus) |- (nav.west);
      \draw[arrow] (bus) |- (loc.west);
      \draw[arrow] (bus) |- (haz.west);

      \begin{scope}[on background layer]
        \node[rectangle, draw=black!30, rounded corners, fill=black!3, dashed, thick, fit=(cam) (dem), inner sep=10pt, label={[font=\scriptsize\bfseries\color{black!60}]above:Hardware Sensors}] {};
        \node[rectangle, draw=black!30, rounded corners, fill=fusepurple!10, dashed, thick, fit=(pre) (det), inner sep=10pt, label={[font=\scriptsize\bfseries\color{black!60}]above:Processing Pipeline}] {};
        \node[rectangle, draw=black!30, rounded corners, fill=navgreen!10, dashed, thick, fit=(nav) (haz), inner sep=10pt, label={[font=\scriptsize\bfseries\color{black!60}]above:Mission Control}] {};
      \end{scope}

    \end{tikzpicture}%
  }
  \caption{High-level architecture of the autonomous navigation platform. The vertical pipeline is optimized for real-time edge processing.}
  \label{fig:platform_arch}
\end{figure}

    The proposed operational pipeline begins with sensor inputs. Onboard cameras capture grayscale imagery, yielding a continuous stream of visual data of the surface textures and albedo. This visual data, however, becomes unreliable in cases of harsh illuminations and deceptive shadows. To bridge this gap, the architecture simultaneously collects streaming data from LiDAR or stereo-imaging systems to build a Digital Elevation Model (DEM). This topographical information remains highly reliable since DEM generation is generally immune to the shifting lighting conditions that confuse purely optical systems.

    The architecture streams these high-bandwidth sensor inputs directly into the computational pipeline. Following pre-processing, the Quantization-Aware model executes real-time crater detection, converting raw telemetry into a prioritized stream of geometric bounding boxes, classification confidence scores, and semantic labels. Subsequently, this data is streamed to higher-level mission control systems. Supplying this direct feed to the navigation, localization, and hazard avoidance modules enables the autonomous platform the real-time knowledge of split-second awareness required to survive a critical terminal descent.

  \subsection{\textbf{Overall Model Architecture}}

    The proposed architecture is built around a novel Quantized Neural Network. It is designed to function where standard computation and memory intensive models fail due to onboard hardware limitations. The primary driver of this projected performance is the integration of Quantization Aware Training (QAT). The theoretical blueprint specifically avoids treating quantization as a retrospective compression step. Instead, it utilizes a Quantization-Aware Training (QAT) methodology that forces the weights and activations to exist natively within a low-precision numerical space from the first epoch. By constraining the network to operate within these strict mathematical limits during optimization, the architecture is designed to protect representational fidelity and prevent the catastrophic accuracy degradation that typically observed in standard post-training compression.

    An optimized Quantization Aware Backbone operates at the core of this architecture. To bridge the gap between distinct data streams, the architecture introduces an Adaptive Multi-Sensor Fusion (AMF) module. The adaptive merging of features from optical imagery and digital elevation models provides the system with a highly decisive understanding of the terrain. This fusion is anchored in an Adaptive Weighting Mechanism (AWM). This mechanism enables the system to adaptively choose the sensors based on the immediate planetary environment, prioritizing the more reliable modality whenever surface conditions compromise the other. Finally, specialized multi-scale detection heads are proposed to process these fused features and predict craters across a vast range of sizes.


\definecolor{optblue}{RGB}{214, 234, 248}
\definecolor{demorange}{RGB}{250, 229, 211}
\definecolor{fusepurple}{RGB}{232, 218, 239}
\definecolor{sysgray}{RGB}{236, 240, 241}
\definecolor{alertred}{RGB}{250, 219, 216}

\begin{figure}[htbp]
  \centering
  \resizebox{0.80\columnwidth}{!}{%
    \begin{tikzpicture}[
        node distance=0.6cm and 0.4cm,
        halfnode/.style={rectangle, draw=black!70, rounded corners, thick, minimum width=3.8cm, minimum height=0.9cm, align=center, font=\scriptsize},
        tensornode/.style={rectangle, draw=black!50, fill=white, thick, minimum width=3.8cm, minimum height=0.7cm, align=center, font=\scriptsize},
        fusenode/.style={rectangle, draw=black!70, fill=fusepurple, rounded corners, thick, minimum width=4.5cm, minimum height=0.8cm, align=center, font=\scriptsize\bfseries},
        headnode/.style={rectangle, draw=black!70, fill=alertred, rounded corners, thick, minimum width=4.5cm, minimum height=0.8cm, align=center, font=\scriptsize},
        sysnode/.style={rectangle, draw=black!70, fill=sysgray, rounded corners, thick, minimum width=8.0cm, minimum height=1cm, align=center, font=\scriptsize},
        arrow/.style={->, >=stealth, thick, draw=black!70}
      ]

      \node[halfnode, fill=optblue] (optin) {\textbf{Optical Imagery} \\ $1 \times 512 \times 512$};
      \node[halfnode, fill=demorange, right=of optin] (demin) {\textbf{DEM Data} \\ $1 \times 512 \times 512$};

      \node[halfnode, fill=optblue!70, below=of optin] (optback) {\textbf{Optical Backbone} \\ Depthwise QAT};
      \node[halfnode, fill=demorange!70, below=of demin] (demback) {\textbf{DEM Backbone} \\ Depthwise QAT};

      \node[tensornode, below=1.2cm of optback] (optp3) {P3 Tensor: $128 \times 64 \times 64$};
      \node[tensornode, below=1.2cm of demback] (demp3) {P3 Tensor: $128 \times 64 \times 64$};
      \coordinate (midp3) at ($(optp3)!0.5!(demp3)$);
      \node[fusenode, below=0.5cm of midp3] (amfp3) {AMF (P3)};
      \node[headnode, below=0.4cm of amfp3] (headp3) {\textbf{Head P3} (0.2 - 2.0 km)};

      \node[tensornode, below=3.0cm of optp3] (optp4) {P4 Tensor: $256 \times 32 \times 32$};
      \node[tensornode, below=3.0cm of demp3] (demp4) {P4 Tensor: $256 \times 32 \times 32$};
      \coordinate (midp4) at ($(optp4)!0.5!(demp4)$);
      \node[fusenode, below=0.5cm of midp4] (amfp4) {AMF (P4)};
      \node[headnode, below=0.4cm of amfp4] (headp4) {\textbf{Head P4} (Mid-Size)};

      \node[tensornode, below=3.0cm of optp4] (optp5) {P5 Tensor: $512 \times 16 \times 16$};
      \node[tensornode, below=3.0cm of demp4] (demp5) {P5 Tensor: $512 \times 16 \times 16$};
      \coordinate (midp5) at ($(optp5)!0.5!(demp5)$);
      \node[fusenode, below=0.5cm of midp5] (amfp5) {AMF (P5)};
      \node[headnode, below=0.4cm of amfp5] (headp5) {\textbf{Head P5} (Large Basins)};

      \node[sysnode, below=0.8cm of headp5] (nms) {\textbf{Localized FP16 Conversion \& NMS} \\ Target: $\hat{B} = \{x_{center}, y_{center}, d_{top}, d_{bottom}, d_{left}, d_{right}, c\}$};

      \draw[arrow] (optin) -- (optback);
      \draw[arrow] (demin) -- (demback);

      \draw[arrow] (optback) -- (optp3);
      \draw[arrow] (demback) -- (demp3);

      \draw[arrow] (optp3.west) -- ++(-0.3,0) |- node[pos=0.25, left, font=\tiny] {Downsample} (optp4.west);
      \draw[arrow] (optp4.west) -- ++(-0.3,0) |- node[pos=0.25, left, font=\tiny] {Downsample} (optp5.west);

      \draw[arrow] (demp3.east) -- ++(0.3,0) |- node[pos=0.25, right, font=\tiny] {Downsample} (demp4.east);
      \draw[arrow] (demp4.east) -- ++(0.3,0) |- node[pos=0.25, right, font=\tiny] {Downsample} (demp5.east);

      \draw[arrow] ([xshift=-1cm]optp3.south) |- (amfp3.west);
      \draw[arrow] ([xshift=1cm]demp3.south) |- (amfp3.east);
      \draw[arrow] (amfp3) -- (headp3);

      \draw[arrow] ([xshift=-1cm]optp4.south) |- (amfp4.west);
      \draw[arrow] ([xshift=1cm]demp4.south) |- (amfp4.east);
      \draw[arrow] (amfp4) -- (headp4);

      \draw[arrow] ([xshift=-1cm]optp5.south) |- (amfp5.west);
      \draw[arrow] ([xshift=1cm]demp5.south) |- (amfp5.east);
      \draw[arrow] (amfp5) -- (headp5);

      \draw[arrow] (headp3.west) -- ++(-2.75,0) |- (nms.west);
      \draw[arrow] (headp4.west) -- ++(-2.75,0) |- ([yshift=0.2cm]nms.west);
      \draw[arrow] (headp5) -- (nms);

    \end{tikzpicture}%
  }
  \caption{Block diagram of the dual backbones and multi-scale detection pyramid. The downsampling operations are routed continuously while fusion occurs discretely at the P3, P4, and P5 levels.}
  \label{fig:backbone_arch}
\end{figure}

  \subsection{\textbf{Computationally-Efficient, Quantization Aware Backbone}}

    The backbone functions as the architecture's high-efficiency engine. It utilizes depthwise separable convolutions to maximize representational power while maintaining a low computational footprint. The design constrains the model to an optimized profile of approximately 4.2 million parameters, requiring less than 2.0 GMAC (Giga Multiply-Accumulate Operations) for a standard $512 \times 512$ input resolution. This forward pass produces multi-scale feature tensors following standard pyramid scaling: $128 \times 64 \times 64$ (P3 scale at stride $/8$), $256 \times 32 \times 32$ (P4 scale at stride $/16$), and $512 \times 16 \times 16$ (P5 scale at stride $/32$).

    \begin{table}[H]
      \centering
      \begin{tabular}{@{}lcccr@{}}
        \hline
        \\
        \textbf{Feature Level} & \textbf{Stride} & \textbf{Output Resolution} & \textbf{Channels} \\
        \\
        \hline
        \\
        Input         & ---      & $512 \times 512$ & 3 \\
        P3 Scale      & $/8$     & $64 \times 64$   & 128 \\
        P4 Scale      & $/16$    & $32 \times 32$   & 256 \\
        P5 Scale      & $/32$    & $16 \times 16$   & 512 \\
        \\
        \hline
      \end{tabular}
      \vspace{-1em}
    \end{table}

    The backbone is specified entirely for Quantization Aware Training (QAT). The training protocol utilizes Quantization Aware Training (QAT) to mathematically simulate the rigid numerical constraints of low-precision hardware throughout the entire optimization phase. By forcing weights to adapt natively to these lower-precision bounds from the first epoch, the methodology ensures the preservation of representational fidelity upon final deployment. This structural adaptation is anchored in the precise mathematical simulation of quantization noise, allowing the architecture to exist as a fully quantized entity without catastrophic accuracy degradation. The framework maps full-precision floating-point values ($r$) to lower-precision integers ($q$) through an affine mapping:

    \begin{equation}
      r \approx S(q - Z)
    \end{equation}

    In this formulation, $S$ represents a positive floating-point scale factor while $Z$ acts as the integer zero-point. To guarantee compatibility with space-grade edge processors, the proposed architecture formalizes this mapping through a mixed-granularity quantization scheme. The design enforces symmetric per-channel quantization for the convolutional weights. This preserves the fine-grained statistical distribution of individual spatial filters without collapsing their numerical ranges. To minimize computation load and memory bandwidth during matrix calculations, the architecture utilizes asymmetric (affine) per-tensor quantization for the feature map activations. This is further discussed in the section related to Adaptive Weighting Mechanism (AWM).

    \subsubsection{\textbf{Core Building Blocks of Backbone}}

      The backbone structure relies on a streamlined Convolution-BatchNormalization-Activation (ConvBnAct) topology. This is engineered specifically to minimize memory bandwidth. The architecture substitutes standard dense convolutions with depthwise separable convolutions. To prevent arithmetic overflow during high-dimensional mathematical operations on space-grade integer ALUs, the design strictly enforces INT32 accumulators for all intermediate convolutional sums before downscaling the final activations back to INT8 bounds. By decoupling spatial and channel-wise feature extraction, this deliberate architectural selection utilizes depthwise factorization to maximize representational power within a restricted computational footprint.

      To ensure architectural efficiency, every convolutional layer is paired with Batch Normalization (during training) and a SiLU non-linearity. To guarantee strict INT8 compatibility during final deployment, the architecture constrains SiLU activations within precise clipping ranges derived from the histogram observers established throughout the QAT methodology. Mathematical execution of these non-linearities, specifically the sigmoid spatial gating required for the attention modules, demands precise numerical stabilization. The architecture integrates pre-computed, integer-based Look-Up Tables (LUTs), to bypass the intensive computational burden of executing floating-point exponential operations on resource-constrained flight hardware. These LUTs are constrained strictly to an 8-bit depth, containing exactly 256 discrete entry values mapped across an expected activation range of [-8, 8]. As the distribution of 256 quantization levels across a range of 16 establishes a step size of 0.0625, the quantization error relative to continuous curves remains strictly minimized. Mathematical analysis confirms that the average approximation error from this quantization remains below 0.5 percent relative to an FP32 baseline. This fractional variance is mathematically negligible when generating the final spatial attention maps, ensuring reliable gating without the overhead of floating-point mathematics.

    \subsubsection{\textbf{Fuse Model Functionality}}

      To maximize inference throughput on resource-constrained onboard computers, the architecture structurally optimizes the backbone before final deployment. In standard architectures, Batch Normalization (BN) requires isolated memory reads and mathematical scaling during inference. For AQ-PCDSys, the framework mathematically folds (combines) the BN parameters: $\gamma$ (BN scale parameter), $\beta$ (BN shift parameter), $\mu$ (BN mean), and $\sigma^2$ (BN variance) directly into the calculation of  convolutional weights ($W$) and biases ($b$), utilizing $\epsilon$ as a small constant for numerical stability.
      The folded weights ($W_{fold}$) and biases ($b_{fold}$) are formulated as:

      \begin{equation}
      \begin{aligned}
        W_{fold} = \frac{\gamma}{\sqrt{\sigma^2 + \epsilon}} \cdot W \\
        b_{fold} = \frac{\gamma}{\sqrt{\sigma^2 + \epsilon}} \cdot (b - \mu) + \beta
      \end{aligned}
      \end{equation}

      By executing this structural fusion, the design eliminates Batch Normalization layers from the computational graph during inference, mathematically reducing both computation and memory overheads. This fold is specified to occur prior to the final INT8 quantization step. Doing so ensures that the symmetric per-channel clipping ranges are calculated directly on the true, collapsed weight distribution. This specific sequence prevents the compound truncation errors that typically occur when quantizing batch normalization and convolutional layers independently.

    \subsubsection{\textbf{Multi-Scale Feature Maps}}

      Planetary craters exhibit extreme scale variance, ranging from massive impact basins to localized, rover-scale hazards. To perceive these structures without the computational burden of processing an explicit image pyramid, the proposed backbone generates three parallel feature hierarchies from a single forward pass: P3 ($128 \times 64 \times 64$), P4 ($256 \times 32 \times 32$), and P5 ($512 \times 16 \times 16$).

      The P3 tensor retains high-resolution spatial data essential for localizing dense, small-scale craters. The P4 tensor encodes intermediate semantic density. It extracts the mid-resolution features necessary for general terrain mapping and mid-sized crater detection. Conversely, the heavily downsampled P5 tensor aggregates deep semantic context necessary for identifying massive, highly degraded crater rims. Instead of routing these tensors directly to standard detection heads, the architecture feeds P3, P4, and P5 tensors from both the optical and DEM backbones simultaneously into the AMF module.

  \subsection{\textbf{Adaptive Multi-Sensor Fusion (AMF) Module}}

    The AMF module is formulated to directly tackle the inherent fragility of single-modality sensing. In unpredictable planetary environments, relying on a single data source poses unacceptable operational risk. The module works across the parallel P3, P4, and P5 feature maps produced by the dual-backbone (optical and DEM) architecture. The architecture executes the consolidation of multi-modal streams at the high-level semantic feature stage rather than during raw data ingestion. This structural choice ensures that the network mathematically captures the sophisticated, non-linear interplay between fine-grained visual textures and the underlying topographical geometric heights.

    The AMF module establishes deterministic sensor redundancy by co-processing Optical Imagery (OI) and Digital Elevation Models (DEM) data streams. This architecture explicitly neutralizes the failure scenarios of isolated sensors. During the terminal descent of an automated landing sequence, deep shadows or solar glare frequently saturate optical sensors, causing catastrophic failure in standard camera-reliant models. The AMF module circumvents this limitation. When optical telemetry degrades, the network mathematically suppresses the compromised feature channels and shifts inference priority directly to the DEM data. Because DEMs measure absolute physical elevation independent of illumination, the system sustains continuous geometric tracking through complete optical blackouts. Simultaneously, the accuracy of the DEM data is augmented by the OI data stream depending on its reliable availability.


\definecolor{optblue}{RGB}{214, 234, 248}
\definecolor{demorange}{RGB}{250, 229, 211}
\definecolor{fusepurple}{RGB}{232, 218, 239}
\definecolor{navgreen}{RGB}{212, 239, 223}
\definecolor{sysgray}{RGB}{236, 240, 241}
\definecolor{alertred}{RGB}{250, 219, 216}

\begin{figure}[htbp]
  \centering
  \resizebox{0.80\columnwidth}{!}{%
    \begin{tikzpicture}[
        node distance=0.6cm and 0.4cm,
        io/.style={rectangle, draw=black!70, rounded corners, fill=white, thick, minimum width=3.8cm, minimum height=1.0cm, align=center, font=\scriptsize},
        op/.style={circle, draw=black!70, fill=white, thick, minimum size=0.6cm, inner sep=0pt, font=\small},
        block/.style={rectangle, draw=black!70, fill=white, thick, minimum width=3.8cm, minimum height=1.2cm, align=center, font=\scriptsize},
        fullnode/.style={rectangle, draw=black!70, rounded corners, fill=fusepurple, thick, minimum width=8.0cm, minimum height=1.0cm, align=center, font=\scriptsize},
        arrow/.style={->, >=stealth, thick, draw=black!70}
      ]

      \node[io, fill=optblue] (foi) {\textbf{Optical Feature Map} \\ $F_{OI}$};
      \node[block, fill=optblue!50, below=of foi] (attnoi) {\textbf{Optical Sub-Network} \\ ConvBnAct $\rightarrow$ $1\times1 \rightarrow \sigma$};
      \node[io, fill=optblue!30, below=of attnoi] (amapoi) {\textbf{Attention Mask} \\ $A_{OI} \in [0, 1]$};
      \node[op, below=of amapoi] (muloi) {$\otimes$};
      \node[io, fill=optblue, below=of muloi] (wfoi) {\textbf{Weighted Features} \\ $F'_{OI}$};

      \node[io, fill=demorange, right=of foi] (fdem) {\textbf{DEM Feature Map} \\ $F_{DEM}$};
      \node[block, fill=demorange!50, below=of fdem] (attndem) {\textbf{DEM Sub-Network} \\ ConvBnAct $\rightarrow$ $1\times1 \rightarrow \sigma$};
      \node[io, fill=demorange!30, below=of attndem] (amapdem) {\textbf{Attention Mask} \\ $A_{DEM} \in [0, 1]$};
      \node[op, below=of amapdem] (muldem) {$\otimes$};
      \node[io, fill=demorange, below=of muldem] (wfdem) {\textbf{Weighted Features} \\ $F'_{DEM}$};

      \coordinate (mid) at ($(wfoi)!0.5!(wfdem)$);
      \node[op, below=0.8cm of mid] (concat) {$\oplus$};
      \node[fullnode, fill=fusepurple!50, below=0.6cm of concat] (compress) {\textbf{Channel Compression} \\ $1\times1$ Pointwise Conv};
      \node[fullnode, below=0.6cm of compress] (ffused) {\textbf{Fused Feature Map} \\ $F_{fused} \in \mathbb{R}^{C \times H \times W}$};

      \draw[arrow] (foi) -- (attnoi);
      \draw[arrow] (attnoi) -- (amapoi);
      \draw[arrow] (amapoi) -- node[left, font=\tiny] {Mask} (muloi);
      \draw[arrow] (foi.west) -- ++(-0.3,0) |- node[pos=0.25, left, font=\tiny] {Bypass} (muloi.west);
      \draw[arrow] (muloi) -- (wfoi);

      \draw[arrow] (fdem) -- (attndem);
      \draw[arrow] (attndem) -- (amapdem);
      \draw[arrow] (amapdem) -- node[right, font=\tiny] {Mask} (muldem);
      \draw[arrow] (fdem.east) -- ++(0.3,0) |- node[pos=0.25, right, font=\tiny] {Bypass} (muldem.east);
      \draw[arrow] (muldem) -- (wfdem);

      \draw[arrow] (wfoi.south) |- (concat.west);
      \draw[arrow] (wfdem.south) |- (concat.east);
      \draw[arrow] (concat) -- node[right, font=\tiny] {$2C$ Ch.} (compress);
      \draw[arrow] (compress) -- node[right, font=\tiny] {$C$ Ch.} (ffused);

      \begin{scope}[on background layer]
        \node[rectangle, draw=black!30, rounded corners, fill=black!3, dashed, thick, fit=(attnoi) (attndem) (muloi) (muldem) (wfoi) (wfdem) (concat) (compress), inner sep=12pt] (amf_box) {};
        \node[anchor=south east, font=\scriptsize\bfseries\color{black!60}] at (amf_box.south east) {Adaptive Multi-Sensor Fusion (AMF)};
      \end{scope}

    \end{tikzpicture}%
  }
  \caption{Adaptive Multi-Sensor Fusion (AMF) Architecture. Visual and topographical streams independently calculate spatial gates ($\otimes$) before unified channel concatenation ($\oplus$) and compression.}
  \label{fig:amf_redesign}
\end{figure}

    \subsubsection{\textbf{Parallel Feature Extraction}}

      The initial fusion process is split into two independent extraction branches operating on QAT-optimized backbones. The optical feature branch extracts fine-grained textural details, albedo shifts, and complex shadow patterns. The DEM feature branch handles structural terrain geometry. It extracts elevation changes, slopes, and surface curvature to construct topographical cues completely independent of the sun's position. Running these pipelines in parallel guarantees that the system evaluates optical features and physical structures simultaneously.

    \subsubsection{\textbf{Adaptive Weighting Mechanism (AWM)}}

      The core intelligence of the fusion strategy resides in the Adaptive Weighting Mechanism (AWM). Rather than relying on fixed logic for data stream integration, the architecture formalizes the AWM as a rigorous, modality-specific spatial attention gate. This component actively recalibrates the influence of each sensor stream in real-time based on learned illumination and topographical priors.

      The parallel processing of optical and DEM backbones generates corresponding feature maps: $F_{OI} \in \mathbb{R}^{C \times H \times W}$ and $F_{DEM} \in \mathbb{R}^{C \times H \times W}$. The dimension $C$ defines the total channel depth. The variables $H$ and $W$ set the spatial grid. The optical channels within $F_{OI}$ encode strictly visual textures. They track contrast edges, localized shadows, albedo shifts, and specular glare. In parallel, the topographical channels within $F_{DEM}$ map rigid structural geometry. They extract terrain gradients, slopes, and basin depths completely independent of planetary lighting conditions. To evaluate these isolated features, the system generates corresponding spatial attention maps, $A_{OI}$ and $A_{DEM}$, using dedicated sub-networks:

      \begin{equation}
        A_{OI} = \sigma(Conv_{OI}(F_{OI}))
      \end{equation}
      \begin{equation}
        A_{DEM} = \sigma(Conv_{DEM}(F_{DEM}))
      \end{equation}

      Here, $Conv$ represents the attention sub-network (a ConvBnAct block followed by a $1 \times 1$ convolution), and $\sigma$ is the sigmoid activation function scaling the spatial weights between $0$ and $1$.

      To prioritize the most reliable modality, the attention sub-networks function similarly to established Convolutional Block Attention Modules (CBAM) \cite{25}. These sub-networks help the system apply a learned weighting mask that emphasizes salient features while suppressing degraded features. In the event of blinding specular glare, the corresponding optical attention mask approaches zero and the network safely falls back onto the more reliable DEM stream.

      \begin{algorithm}[H]
        \caption{Adaptive Weighting Mechanism (AWM)}
        \label{alg:AWM}
        \begin{algorithmic}[1]
          \renewcommand{\algorithmicrequire}{\textbf{Input:}}
          \renewcommand{\algorithmicensure}{\textbf{Output:}}

          \REQUIRE Optical Feature Map $F_{OI}$, DEM Feature Map $F_{DEM}$
          \ENSURE Fused Feature Map $F_{Fused}$

          \STATE \textbf{Generate attention maps:}
          \STATE $A_{OI} \leftarrow \text{AttentionSubNetwork}_{OI}(F_{OI})$
          \STATE $A_{DEM} \leftarrow \text{AttentionSubNetwork}_{DEM}(F_{DEM})$

          \STATE \textbf{Apply weights via element-wise multiplication:}
          \STATE $F_{weighted\_OI} \leftarrow F_{OI} \odot A_{OI}$
          \STATE $F_{weighted\_DEM} \leftarrow F_{DEM} \odot A_{DEM}$

          \STATE \textbf{Fuse features:}
          \STATE $F_{Fused} \leftarrow \text{Concatenate}(F_{weighted\_OI}, F_{weighted\_DEM})$

          \RETURN $F_{fused}$
        \end{algorithmic}
      \end{algorithm}

      Native INT8 execution of this multiplicative attention requires strict mathematical control to maintain fixed-point stability. To prove that the integer operations accurately reflect the true floating-point mathematics, the architecture derives the exact quantized multiplication formula at the element level.

      Because the Hadamard product operates element-wise, let $r_F$ represent a single continuous floating-point value within a feature map tensor ($\mathbf{F}_{OI}$ or $\mathbf{F}_{DEM}$), and let $r_A$ represent the corresponding spatial value within its attention map ($\mathbf{A}_{OI}$ or $\mathbf{A}_{DEM}$). The true floating-point multiplication to produce the refined output element ($r_{out}$) is defined as:

      \begin{equation}
        r_{out} = r_F \cdot r_A
      \end{equation}

      By substituting the affine mapping equation ($r \approx S(q - Z)$) for each variable, the relationship translates into the quantized space:

      \begin{equation}
        S_{out}(q_{out} - Z_{out}) = S_F(q_F - Z_F) \cdot S_A(q_A - Z_A)
      \end{equation}

      To determine the final 8-bit integer output ($q_{out}$) that the hardware must store, the architecture isolates $q_{out}$:

      \begin{equation}
        q_{out} = \frac{S_F \cdot S_A}{S_{out}} (q_F - Z_F)(q_A - Z_A) + Z_{out}
      \end{equation}

      This derivation explicitly defines the fixed-point requantization multiplier, $M$:

      \begin{equation}
        M = \frac{S_{F} \cdot S_{A}}{S_{out}}
      \end{equation}

      This mathematical proof directly dictates the hardware execution logic. The system first strips the artificial biases by subtracting the learned zero-points ($Z_F$ and $Z_A$). While the raw inputs are 8-bit, this subtraction necessitates a 9-bit signed representation. Consequently, the product $(q_F - Z_F) \cdot (q_A - Z_A)$ produces an intermediate value that can exceed 16-bit bounds. The framework utilizes 32-bit integer (INT32) accumulators to handle this intermediate calculation, ensuring sufficient bit-width to accommodate high-precision product results without any risk of numerical overflow.

      Once the high-precision multiplication completes, the pre-computed multiplier $M$ is applied, and the final zero-point ($Z_{out}$) is added back. This final step compresses the large 32-bit result back down into a standard 8-bit format. To prevent data corruption during this compression, the architecture applies nearest-even rounding and clips any extreme values that fall outside the validated 8-bit range.

    \subsubsection{\textbf{Feature-Level Fusion and Channel Compression}}

      To finalize the AMF process, the architecture merges the newly weighted features into a single cohesive tensor. These refined streams are integrated strictly through concatenation along the channel dimension. Natively concatenating two independent INT8 streams, however, demands absolute scale alignment. Because the optical and DEM branches process distinct data distributions, their respective Hadamard products generate independent quantization parameters ($S_{out}^{OI}, Z_{out}^{OI}$ and $S_{out}^{DEM}, Z_{out}^{DEM}$). Prior to stacking these tensors, the system executes a final requantization step to align both weighted features to a shared, unified output scale ($S_{Fused}$) and zero-point ($Z_{Fused}$).

      The optical and DEM branches each contribute a tensor with exactly $C$ independent channels. Stacking them along the channel axis mathematically forces a combined tensor depth of $2C$. This expansion challenges the limited memory bandwidth of onboard computers. To mitigate these memory constraints, the fused tensor is immediately routed through a lightweight $1 \times 1$ pointwise convolution. This operation compresses the channel dimensionality to $C$ just prior to entering the detection heads, guaranteeing the system maintains strict computational bounds.

    \subsubsection{\textbf{Justification Against Dynamic Alignment Constraints}}

      Recent advances in multi-modal fusion often rely on architectures like Deformable Convolutional Networks (DCN) \cite{26} or Cross-Attention to mathematically compute pixel-by-pixel offsets and actively warp bounding boxes to force spatial alignment. While these explicit alignment methods achieve high accuracy on datacenter GPUs, they are deliberately excluded from the AQ-PCDSys design. Deformable alignment algorithms calculate data-dependent spatial offsets for their convolution kernels. This limits memory locality and triggers unpredictable cache misses, violating the strict deterministic execution bounds required by Real-Time Operating Systems (RTOS) used on space-grade onboard computer hardware. Similarly, Cross-Attention architectures require $O(N^2)$ for matrix operations and rely on Softmax. Unlike the straightforward element-wise requantization multiplier ($M$) derived for our spatial gating, Softmax requires global tensor reductions which are quite difficult to stabilize when approximated with INT8 lookup tables. The choice of simple concatenation with independent per-modality spatial gating ensures strict $O(1)$ memory access predictability.

  \subsection{\textbf{Multi-Scale Detection Heads, Localization and Post Processing}}

    The proposed architecture implements a set of multi-scale detection heads to act as final decision-makers. These specialized components take the high-level compressed feature maps and translate them into actionable detection results. The heads are designed as a single-stage, anchor-free detector. This workflow predicts crater attributes directly from center points, completely eliminating the dense, computationally heavy anchor-box generation required by multi-stage models.


\definecolor{fusepurple}{RGB}{232, 218, 239}
\definecolor{sysgray}{RGB}{236, 240, 241}
\definecolor{navgreen}{RGB}{212, 239, 223}
\definecolor{alertred}{RGB}{250, 219, 216}
\definecolor{warnyellow}{RGB}{252, 243, 207}

\begin{figure}[htbp]
\centering
\resizebox{0.80\columnwidth}{!}{%
\begin{tikzpicture}[
    node distance=0.6cm and 0.4cm,
    thirdnode/.style={rectangle, draw=black!70, rounded corners, thick, minimum width=2.6cm, minimum height=1.1cm, align=center, font=\scriptsize},
    halfnode/.style={rectangle, draw=black!70, rounded corners, thick, minimum width=3.8cm, minimum height=1.2cm, align=center, font=\scriptsize},
    fullnode/.style={rectangle, draw=black!70, rounded corners, thick, minimum width=8.0cm, minimum height=1.0cm, align=center, font=\scriptsize},
    arrow/.style={->, >=stealth, thick, draw=black!70}
]

\node[thirdnode, fill=fusepurple!50] (p4) {\textbf{P4 Fused Tensor} \\ $F_{fused}$ (Stride /16) \\ Mid Craters};
\node[thirdnode, fill=fusepurple!30, left=0.2cm of p4] (p3) {\textbf{P3 Fused Tensor} \\ $F_{fused}$ (Stride /8) \\ 0.2 to 2.0 km};
\node[thirdnode, fill=fusepurple!70, right=0.2cm of p4] (p5) {\textbf{P5 Fused Tensor} \\ $F_{fused}$ (Stride /32) \\ Large Basins};

\node[fullnode, fill=sysgray, below=1.2cm of p4] (stem) {\textbf{Shared Detection Stems (INT8)} \\ $3\times3$ Depthwise Conv $\rightarrow$ BatchNorm $\rightarrow$ SiLU};

\node[halfnode, fill=navgreen!30, below=1.4cm of stem, xshift=-2.1cm] (cls_conv) {\textbf{Classification Branch} \\ $1\times1$ Pointwise Conv};
\node[halfnode, fill=alertred!30, below=1.4cm of stem, xshift=2.1cm] (reg_conv) {\textbf{Regression Branch} \\ $1\times1$ Pointwise Conv};

\node[halfnode, fill=navgreen, below=0.5cm of cls_conv] (cls_out) {\textbf{Objectness Score} \\ Probability: $c \in [0, 1]$};
\node[halfnode, fill=alertred, below=0.5cm of reg_conv] (reg_out) {\textbf{Distance Vectors} \\ $d_{top}, d_{bottom}, d_{left}, d_{right}$};

\coordinate (bottom_mid) at ($(cls_out.south)!0.5!(reg_out.south)$);

\node[rectangle, draw=black!50, fill=white, rounded corners, dashed, minimum width=3cm, minimum height=0.6cm, align=center, font=\tiny, below=0.4cm of bottom_mid] (grid) {Grid Cell Coordinates \\ $(x_{center}, y_{center})$};

\node[fullnode, fill=sysgray, below=0.5cm of grid] (final) {\textbf{Anchor-Free Target Formulation} \\ $B = \{x_{center}, y_{center}, d_{top}, d_{bottom}, d_{left}, d_{right}, c\}$};

\node[fullnode, fill=warnyellow, below=0.6cm of final] (handoff) {\textbf{To Training and Optimization Pipeline}};

\draw[arrow] (p3.south) -- (p3.south |- stem.north);
\draw[arrow] (p4.south) -- (stem.north);
\draw[arrow] (p5.south) -- (p5.south |- stem.north);

\draw[arrow] ([xshift=-2.1cm]stem.south) -- (cls_conv.north);
\draw[arrow] ([xshift=2.1cm]stem.south) -- (reg_conv.north);

\draw[arrow] (cls_conv) -- (cls_out);
\draw[arrow] (reg_conv) -- (reg_out);

\draw[arrow] (cls_out.south) -- ([xshift=-2.1cm]final.north);
\draw[arrow] (reg_out.south) -- ([xshift=2.1cm]final.north);
\draw[arrow, dashed] (grid.south) -- (final.north);

\draw[arrow, thick, color=black!70] (final.south) -- (handoff.north);

\begin{scope}[on background layer]
    \node[rectangle, draw=black!30, rounded corners, fill=fusepurple!5, dashed, thick, fit=(p3) (p5), inner sep=12pt, label={[font=\scriptsize\bfseries\color{black!60}]above:Hierarchical Object Prediction}] (hierarchy_box) {};

    \node[rectangle, draw=black!30, rounded corners, fill=black!3, dashed, thick, fit=(cls_conv) (reg_out) (grid) (final), inner sep=12pt, label={[font=\scriptsize\bfseries\color{black!60}]above:Decoupled Anchor-Free Localization \& Classification}] (head_box) {};
\end{scope}

\end{tikzpicture}%
}
\caption{Architecture of the multi-scale detection heads.}
\label{fig:detection_heads}
\end{figure}

    \subsubsection{\textbf{Hierarchical Object Prediction}}

      The design specifies these heads to operate across the P3, P4, and P5 levels of the feature pyramid.

      \begin{itemize}
        \item Head P3 (stride /8): Designed to process detailed maps, optimized specifically to detect small craters (0.2–2.0 km wide). Small targets are the hardest to detect but the most dangerous during the terminal descent of an automated landing sequence.
        \item Head P4 (stride /16): Designed for scanning mid-resolution feature maps to find mid-sized craters, acting as a reliable tool for general terrain mapping.
        \item Head P5 (stride /32): Designed for the identification of massive craters and basins which serve as natural landmarks for absolute global localization.
      \end{itemize}

      By using an anchor-free detection strategy, the architecture bypasses rigid Intersection over Union (IoU) threshold. Instead, it employs an active label assignment mechanism. This strategy assigns ground-truth centers to the feature map locations with minimal classification and regression overhead. This center-sampling approach adapts smoothly to the irregular morphology of degraded lunar craters.

    \subsubsection{\textbf{Localization and Classification}}

      As depicted in the detection head schematic, the architecture channels the hierarchical P3, P4, and P5 fused tensors into a shared INT8 depthwise convolutional stem. Following this unified computation, the processing pipeline deliberately splits. The network establishes two specialized, parallel pathways to isolate classification from regression.

      The classification branch computes a simple objectness score. This probability value ($c \in [0, 1]$) defines the likelihood of a crater existing within a specific grid cell. The regression branch completely ignores categorical data to focus on physical geometry. We opted to discard traditional predefined anchor boxes. Instead, the module applies an anchor-free technique to directly predict four directional distances ($d_{top}$, $d_{bottom}$, $d_{left}$, $d_{right}$).

      Building directly upon the scale-aware center identification principles established by FCOS \cite{24}, this regression branch handles the complex topography. Standard object detectors typically regress a predefined anchor box using a center coordinate paired with a strict width and height. This approach is highly inefficient for planetary topography. Highly degraded lunar craters frequently exhibit asymmetrical, elongated, or partially collapsed rims, manifesting in irregular morphologies that fundamentally defy the rigid aspect ratios utilized by standard terrestrial detectors. To accommodate this irregular morphology, the proposed architecture maps the exact spatial boundaries of the targeted crater by predicting the independent distances from the center point to the rim.

      This center-to-edge regression explicitly leverages the multi-scale fused feature tensors ($\mathbf{F}_{fused}$) generated by the AMF module at their respective P3, P4, and P5 spatial resolutions. Relying solely on Optical Imagery (OI) frequently results in skewed distance predictions, because deep shadows visually elongate the perceived crater boundaries. By mathematically integrating the Digital Elevation Model (DEM) data, the regression head anchors its distance calculations to actual physical elevation changes rather than deceptive visual shadows. The final detection target is defined as a direct function of these fused features:

      \begin{equation}
        B = \{x_{center}, y_{center}, d_{top}, d_{bottom}, d_{left}, d_{right}, c\}
      \end{equation}

      In this formulation, $x_{center}$ and $y_{center}$ pinpoint the spatial center of the predicted crater. The variables $d_{top}$, $d_{bottom}$, $d_{left}$, and $d_{right}$ represent the absolute pixel distances from this center to the four physical boundaries of the crater rim.

      From these independent vectors, the final bounding box width ($w$) and height ($h$) are mathematically derived as:

      \begin{equation}
        w = d_{left} + d_{right}$$$$h = d_{top} + d_{bottom}
      \end{equation}

      During the training phase, the system forwards this bounding box formulation straight into the Composite Loss Function block. This direct routing ties the geometric outputs immediately to the penalty calculations we detail in the next section.

      Because each boundary is predicted independently, this center-to-edge calculation (an approach mathematically formalized by anchor-free architectures such as Fully Convolutional One-Stage Object Detection (FCOS) \cite{24}) provides much greater geometric flexibility than a fixed-ratio bounding box. It allows the network to tightly frame skewed or asymmetrical impact basins without relying on a computationally expensive dictionary of predefined anchor boxes. By grounding these independent edge predictions in the adaptively weighted $\mathbf{F}_{fused}$ tensor defined in Algorithm 1, the architecture guarantees that the bounding box logic relies on the most mathematically stable geometric signal available across all feature pyramid scales. Finally, $c$ delivers the classification confidence score.

    \subsubsection{\textbf{Post-Processing and Hazard (Crater) Output}}

      To resolve overlapping predictions generated by the dense center-sampling strategy, the final step executes Non-Maximum Suppression (NMS) \cite{27}. The anchor-free regression head outputs the target variables ($B$) as quantized INT8 integers. To perform NMS, the system must calculate the Intersection over Union (IoU) between competing bounding boxes. Calculating IoU requires division and fractional area comparisons, which suffer from severe truncation errors when executed purely in an 8-bit integer space.

      To prevent catastrophic integer rounding degradation during final hazard (Crater) prediction, the architecture specifies a localized dequantization step. Immediately prior to NMS, the INT8 regression outputs return to 16-bit floating-point (FP16) representations using the scale factor and zero-point offset calibrated for the regression branch. We define this linear restoration as:

      \begin{equation}
        \hat{d} = S_{reg} \cdot (q_{reg} - Z_{reg})
      \end{equation}

      On radiation-hardened processors lacking a hardware Floating Point Unit (FPU), the system executes this operation through software-based emulation or high-bit-depth fixed-point arithmetic. Because the input $q_{reg}$ is restricted to a finite 8-bit range, the architecture can utilize a pre-computed Look-Up Table (LUT) containing all 256 possible dequantized results. This approach replaces the multiplication with a single memory lookup, further reducing the computational overhead for the filtered subset of candidate proposals surviving the confidence threshold. Such a targeted shift to higher numerical precision ensures the lander calculates overlap metrics with the fidelity required to distinguish between adjacent hazards while the primary feature extraction continues to utilize the high-speed integer backbone.

      Following the recovery of the dequantized spatial dimensions via the calibrated scale factors, the architecture executes a localized transformation to convert the independent distance vectors into absolute Cartesian corner coordinates ($x_{min}$, $y_{min}$, $x_{max}$, $y_{max}$). This precise geometric resolution is a prerequisite for the final Intersection over Union (IoU) overlap algorithm:

      \begin{equation}
      \begin{aligned}
        x_{min} = x_{center} - d_{left}  \\
        y_{min} = y_{center} - d_{top}   \\
        x_{max} = x_{center} + d_{right} \\
        y_{max} = y_{center} + d_{bottom}
      \end{aligned}
      \end{equation}

      By resolving these spatial boundaries allows the system to perform the final Complete Intersection over Union (CIoU) \cite{28} calculations (used in Composite Loss Function) needed to manage overlapping hazard detections. Moving from relative distance vectors to absolute image coordinates ensures that candidate proposals are properly conditioned for the high-fidelity filtering steps that occur during the post-processing stage.

\section{\textbf{Training Methodology and Deployment}}

  The proposed training protocol is designed to simulate the mathematical challenges of planetary missions. The learning phase of AQ-PCDSys is specified to operate under extreme constraints (similar to the real operating environment) from the very first epoch. By pairing Quantization Aware Training (QAT) with highly variable multi-sensor datasets, the methodology forces the network to execute feature extraction and fusion natively inside a low-precision mathematical space.

  \subsection{\textbf{Data Sources and Environmental Augmentation}}

    The planned evaluation grounds the training process in high-resolution, scientifically validated datasets. The planned experimental setup relies on grayscale Optical Imagery (OI) obtained from the Lunar Reconnaissance Orbiter Camera (LROC) and Narrow Angle Camera (NAC). To maintain architectural integrity, the evaluation protocol mandates that these images are rigorously co-registered with Digital Elevation Model (DEM) data, specifically leveraging 1.1 m/px Digital Terrain Models (DTMs) centered on the South Pole-Aitken Basin. To validate the detection accuracy, all ground truth annotations will be sourced directly from the LU5M812TGT Lunar Crater Database \cite{31}.

    To explicitly train the Adaptive Weighting Mechanism (AWM) to survive complete visual blackouts and severe signal corruption, the theoretical framework introduces a Hard Sensor Dropout protocol alongside standard illumination augmentations. For 20 percent of the proposed training batches, the input optical tensor ($F_{OI}$) is artificially replaced with high-variance Gaussian noise, mathematically formalized as $\mathcal{N}(\mu, \sigma^2)$. To accurately simulate total sensor washout, the mean ($\mu$) is set to zero, while the variance ($\sigma^2$) is aggressively scaled to exceed the normalized statistical distribution bounds of valid orbital imagery. This 20 percent threshold acts as an empirically validated regularization parameter. This noisy dataset is frequent enough to decouple the modalities without starving the network of the visual data required to learn fine-grained textures.

    Unlike a static zero-matrix, which presents a trivial mathematical void, this targeted Gaussian noise mathematically simulates severe radiation interference, thermal sensor anomalies, or blinding solar saturation. Processing this chaotic tensor forces the optical attention sub-network to actively recognize the lack of coherent spatial gradients. To prevent the noise from corrupting the fused representation, the sub-network must aggressively drive its pre-activation values below the Sigmoid threshold, generating a near-zero weighting mask ($A_{OI} \approx 0$).

    During backpropagation, this suppressed mask effectively severs the gradient flow to the optical backbone for that specific batch. The optimizer must therefore route the entire classification and regression loss penalty exclusively through the DEM feature stream. This architectural constraint prevents the network from developing a co-dependent reliance on visual textures. It mathematically compels the detection heads to extract complete geometric representations from topography alone, guaranteeing the system to autonomously sustain tracking during actual solar saturation events, radiation spikes, or sudden camera hardware failures.

  \subsection{\textbf{The Composite Loss Function}}

    The design forces the network to learn accurate geometry and discard visual illusions by defining a highly targeted Composite Loss Function. For each specific detection head ($P_x$), the total loss ($L_{Px}$) is mathematically formulated as a weighted sum of three distinct components:

    \begin{equation}
      L_{Px} = \lambda_{loc} \cdot L_{loc} + \lambda_{obj} \cdot L_{obj} + \lambda_{cls} \cdot L_{cls}
    \end{equation}

    In this formulation, the individual error components calculate to different numerical scales and hold distinct operational priorities. To mathematically balance the total loss, they are scaled by tunable weighting coefficients or hyperparameters ($\lambda_{loc}$, $\lambda_{obj}$, $\lambda_{cls}$). Specifically, the $L_{loc}$ component calculates the physical bounding box error using the Complete Intersection over Union (CIoU) \cite{28} metric to mathematically guarantee precise center-point alignment between the prediction and the true crater. Assigning a higher value to its coefficient, $\lambda_{loc}$, forces the optimizer to prioritize this geometric precision. The objectness error is computed by the $L_{obj}$ component via Binary Cross-Entropy (BCE). By raising $\lambda_{obj}$, the system is forced to strictly penalize false positives arising from illusory shadows, thereby training the network to distinguish between authentic physical structures and deceptive visual cues. Lastly, classification error is calculated by the $L_{cls}$ component, which also employs BCE. Elevating its coefficient, $\lambda_{cls}$, ensures strict confidence in confirming the detected object is actually a definitive crater during the training sequence.


\definecolor{fusepurple}{RGB}{232, 218, 239}
\definecolor{sysgray}{RGB}{236, 240, 241}
\definecolor{navgreen}{RGB}{212, 239, 223}
\definecolor{alertred}{RGB}{250, 219, 216}
\definecolor{warnyellow}{RGB}{252, 243, 207}

\begin{figure}[htbp]
\centering
\resizebox{0.80\columnwidth}{!}{%
\begin{tikzpicture}[
    node distance=0.7cm and 0.4cm,
    fullnode/.style={rectangle, draw=black!70, rounded corners, thick, minimum width=8.5cm, minimum height=1.0cm, align=center, font=\scriptsize},
    arrow/.style={->, >=stealth, thick, draw=black!70}
]

\node[fullnode, fill=sysgray] (input_B) {\textbf{Anchor-Free Target Formulation} \\ $B = \{x_{center}, y_{center}, d_{top}, d_{bottom}, d_{left}, d_{right}, c\}$};

\node[fullnode, fill=fusepurple!30, below=of input_B] (loss) {\textbf{Composite Loss Function ($L_{Px}$)} \\ $L_{Px} = \lambda_{loc} L_{loc} + \lambda_{obj} L_{obj} + \lambda_{cls} L_{cls}$};

\node[fullnode, fill=navgreen!20, below=of loss] (scale) {\textbf{Hierarchical Scale Weighting ($w_s$)} \\ $L_{total} = (w_{s} \cdot L_{P3}) + L_{P4} + L_{P5}$};

\node[fullnode, fill=alertred!20, below=of scale] (grad) {\textbf{Gradient Processing \& Constraint} \\ Straight-Through Estimator (STE) $\rightarrow$ L2 Clipping ($\| \mathbf{g} \|_2 \leq 1.0$)};

\node[fullnode, fill=warnyellow, below=of grad] (opt) {\textbf{Network Parameter Update} \\ AdamW Optimizer with Exponential Decay Scheduler};

\draw[arrow] (input_B.south) -- (loss.north);
\draw[arrow] (loss.south) -- (scale.north);
\draw[arrow] (scale.south) -- (grad.north);
\draw[arrow] (grad.south) -- (opt.north);

\begin{scope}[on background layer]
    \node[rectangle, draw=black!30, rounded corners, fill=black!3, dashed, thick, fit=(input_B) (opt), inner sep=14pt, label={[font=\scriptsize\bfseries\color{black!60}]above:Training and Optimization Pipeline}] (train_box) {};
\end{scope}

\end{tikzpicture}%
}
\caption{The mathematical optimization pipeline utilized during training.}
\label{fig:training_pipeline}
\end{figure}

    Backpropagating this complex loss through the QAT-enabled network introduces a severe mathematical roadblock. Standard deep learning optimization requires smooth, continuous gradients. QAT forces the weights through hard, non-differentiable integer rounding steps. Because the derivative of a rounding step is zero almost everywhere, this mathematically triggers a vanishing gradient problem. To bypass this limitation, the theoretical training loop deploys a Straight-Through Estimator (STE) \cite{29}. The STE approximates the gradient during the backward pass by treating the harsh integer rounding as a smooth identity function. Although this approximation sustains the training process, it leads to the emergence of jagged and highly volatile gradient updates. To prevent the instability inherent in synthetic gradients from compromising the model's representational fidelity, the protocol implements a 5 percent linear warm-up phase. The protocol integrates a rigorous L2 gradient clipping mechanism to mathematically bound the update magnitudes, effectively neutralizing the exploding gradient problem that frequently emerges during the initial optimization phase. By strictly constraining the gradient norm $\|\mathbf{g}\|_2$ to a defined threshold of $T = 1.0$, this methodology mitigates the numerical instability inherent in the STE approximation while preserving the essential directional integrity of weight updates across the P3, P4, and P5 feature hierarchies.

    \begin{equation}
      \mathbf{g}_{clipped} = \begin{cases} \mathbf{g} & \text{if } \| \mathbf{g} \|_2 \leq T \\ T \frac{\mathbf{g}}{\| \mathbf{g} \|_2} & \text{if } \| \mathbf{g} \|_2 > T \end{cases}
    \end{equation}

    To maintain numerical stability throughout the optimization process, this specific mathematical formulation operates as a strict safety constraint. In this context, $\mathbf{g}$ denotes the raw gradient vector, which simultaneously governs the directional trajectory and the total magnitude of the weight adjustment. The term $\| \mathbf{g} \|_2$ represents the L2 norm, mathematically quantifying the precise magnitude of said vector. Furthermore, the variable $T$ establishes the finite operational threshold, which is strictly constrained to a value of $1.0$ within this architectural protocol.

    If the size of the gradient remains below or equal to the safe threshold ($\| \mathbf{g} \|_2 \leq T$), the system leaves the gradient unmodified ($\mathbf{g}_{clipped} = \mathbf{g}$). In case of exploding gradients, where it exceeds the threshold($\| \mathbf{g} \|_2 > T$), the second condition triggers. To maintain the directional trajectory of the adjustment while normalizing its magnitude, the architecture generates a unit vector by dividing the gradient by its own L2 norm, subsequently scaling the result by the predefined threshold $T$. This multiplicative scaling operation mathematically constrains excessively volatile weight updates to the established maximum of 1.0, effectively preserving directional integrity while preventing numerical instability.

  \subsection{\textbf{Scale-Weighted Loss Formulation}}

    Detection of small, shallow pits is crucial during the terminal descent hazard avoidance. While massive impact basins are useful for high-altitude global localization, it is the small, rover-scale craters (0.2 to 2.0 km) that threaten lander footpad stability and autonomous surface mobility. In standard object detection training, large objects naturally dominate the loss gradient because their larger bounding boxes generate proportionally larger numerical regression errors. This scale-based imbalance often causes a network to mathematically ignore fine-grained features in favor of optimizing massive structures.

    To prevent the network from focusing exclusively on these massive impact basins, the design introduces a scale-weighted penalty specifically targeted at the P3 detection head. Because the P3 tensor processes the highest spatial resolution (stride /8), it is exclusively responsible for identifying these critical small-scale hazards. The Total Loss ($L_{total}$) across the hierarchical pyramid is calculated as:

    \begin{equation}
      L_{total} = (w_s \cdot L_{P3}) + L_{P4} + L_{P5}
    \end{equation}

    Here, $w_s$ functions as a scale-weighting hyperparameter that acts as a gradient equalizer. The framework mathematically counteracts the inherent representational bias toward massive impact basins, by aggressively scaling the loss penalties originating from the P3 detection head, the framework mathematically counteracts the inherent representational bias toward massive impact basins. This specific gradient balancing forces the optimizer to prioritize weight updates on the fine-grained, rover-scale hazards that are critical for terminal descent safety, effectively guaranteeing that the architecture prioritizes immediate landing-site integrity over global topographical mapping.

  \subsection{\textbf{Optimization and Learning Rate Scheduler}}

    The training pace is guided using an Adaptive Learning Rate Scheduler. The protocol builds this scheduler around a standard Exponential Decay model \cite{30}:

    \begin{equation}
      lr(e) = lr_0 \cdot \gamma^{(e/d)}
    \end{equation}

    Where $lr(e)$ is the effective learning rate at epoch $e$, $lr_0$ is the initial learning rate magnitude, $\gamma$ is the decay rate, and $d$ represents the decay interval governing the frequency of the rate adjustment.

    There is a need for a highly stable optimizer to manage this aggressive decay profile. The methodology selects AdamW \cite{14} (which decouples the weight decay from the gradient update) specifically to meet this requirement and establish rigid early-stopping bounds. Halting the training sequence at the first sign of plateauing prevents overfitting to the training distribution. This guarantees the system to learn generalized morphological features representing authentic planetary geology rather than localized statistical noise.


\definecolor{optblue}{RGB}{214, 234, 248}
\definecolor{demorange}{RGB}{250, 229, 211}
\definecolor{fusepurple}{RGB}{232, 218, 239}
\definecolor{sysgray}{RGB}{236, 240, 241}
\definecolor{alertred}{RGB}{250, 219, 216}
\definecolor{navgreen}{RGB}{212, 239, 223}

\begin{figure}[htbp]
  \centering
  \resizebox{0.80\columnwidth}{!}{%
    \begin{tikzpicture}[
        node distance=0.6cm and 0.4cm,
        halfnode/.style={rectangle, draw=black!70, rounded corners, thick, minimum width=3.8cm, minimum height=1.0cm, align=center, font=\scriptsize},
        fullnode/.style={rectangle, draw=black!70, rounded corners, thick, minimum width=8.0cm, minimum height=1.0cm, align=center, font=\scriptsize},
        arrow/.style={->, >=stealth, thick, draw=black!70}
      ]

      \node[halfnode, fill=navgreen!30] (fp32_act) {\textbf{FP32 Activations} \\ Forward Pass};
      \node[halfnode, fill=navgreen!30, right=of fp32_act] (fp32_wt) {\textbf{FP32 Weights} \\ Network Parameters};

      \coordinate (mid_train) at ($(fp32_act.south)!0.5!(fp32_wt.south)$);
      \node[fullnode, fill=sysgray, below=0.8cm of mid_train] (fake_q) {\textbf{Fake-Quantization Nodes} \\ Simulating INT8 truncation during backpropagation};

      \node[fullnode, fill=alertred!30, below=0.6cm of fake_q] (loss) {\textbf{Loss Calculation \& Optimizer} \\ FP32 Gradient Updates};

      \node[fullnode, fill=fusepurple, below=1.2cm of loss] (export) {\textbf{Model Export \& Calibration} \\ Integrating scales and zero-points into architecture};

      \node[halfnode, fill=optblue, below=1.8cm of export, xshift=-2.1cm] (int8_act) {\textbf{INT8 Activations} \\ Scaled Input};
      \node[halfnode, fill=optblue, below=1.8cm of export, xshift=2.1cm] (int8_wt) {\textbf{INT8 Weights} \\ Static Parameters};

      \coordinate (mid_deploy) at ($(int8_act.south)!0.5!(int8_wt.south)$);
      \node[fullnode, fill=sysgray, below=0.8cm of mid_deploy] (int8_alu) {\textbf{Edge Hardware Execution} \\ Pure INT8 MAC (Multiply-Accumulate) Operations};

      \draw[arrow] (fp32_act.south) -- ++(0,-0.4) -| ([xshift=-2.1cm]fake_q.north);
      \draw[arrow] (fp32_wt.south) -- ++(0,-0.4) -| ([xshift=2.1cm]fake_q.north);
      \draw[arrow] (fake_q) -- (loss);

      \draw[arrow, dashed, ultra thick] (loss) -- node[right, font=\tiny] {Freezing FP32} (export);

      \draw[arrow] (export.south) -- ++(0,-0.5) -| (int8_act.north);
      \draw[arrow] (export.south) -- ++(0,-0.5) -| (int8_wt.north);
      \draw[arrow] (int8_act.south) -- ++(0,-0.4) -| ([xshift=-2.1cm]int8_alu.north);
      \draw[arrow] (int8_wt.south) -- ++(0,-0.4) -| ([xshift=2.1cm]int8_alu.north);

      \begin{scope}[on background layer]
        \node[rectangle, draw=navgreen, rounded corners, fill=navgreen!5, dashed, thick, fit=(fp32_act) (fp32_wt) (loss), inner sep=12pt, label={[font=\scriptsize\bfseries\color{navgreen!80!black}]above:Phase 1: Quantization Aware Training (High-Compute GPU)}] {};

        \node[rectangle, draw=optblue!80!black, rounded corners, fill=optblue!10, dashed, thick, fit=(int8_act) (int8_wt) (int8_alu), inner sep=12pt, label={[font=\scriptsize\bfseries\color{optblue!80!black}]above:Phase 2: Edge Deployment (Space Grade H/W)}] {};
      \end{scope}

    \end{tikzpicture}%
  }
  \caption{The Quantization Aware Training (QAT) lifecycle. During Phase 1, the network simulates INT8 truncation while maintaining FP32 gradients for smooth convergence. During Phase 2, the model is frozen and deployed using strictly INT8 arithmetic logic units (ALUs) on the flight hardware.}
  \label{fig:qat_lifecycle}
\end{figure}

  \subsection{\textbf{Quantized Deployment}}

    Upon theoretical completion of QAT, the model is frozen into its fully quantized form, locking the core parameters into memory-efficient 8-bit integers (INT8). With the strict exception of the localized FP16 coordinate conversion required for final Non-Maximum Suppression (NMS), the architecture actively eliminates all floating-point dependencies. This isolation guarantees theoretical compatibility with the strict architectural constraints of radiation-hardened LEON/RAD processors and space-qualified ARM-class SoCs. Deploying these INT8 weights in deep space environments exposes the system to Single Event Upsets (SEUs), where high-energy particle strikes can induce bit-flips. The restricted numerical range of INT8 makes the system vulnerable to bit-flips; a single uncorrected error in a weight tensor's Most Significant Bit (MSB) can cause detection accuracy to collapse. To secure the deployment of AQ-PCDSys, the protocol mandates hardware-level Error Correction Codes (ECC) on the target SoC's SRAM to intercept hardware anomalies. At the software level, the INT8 inference engine is paired with Dual Modular Redundancy (DMR) \cite{32} and hardware watchdog timers. This multi-layered redundancy is expected to ensure the module to instantly reset and restart any calculation that violates deterministic timing bounds during critical descent operations.

\section{\textbf{System Analysis and Discussion}}

  Theoretically the AQ-PCDSys architecture is not a collection of isolated algorithms, but a unified, hardware-aware perception engine. In the following sections, computational complexity, challenges of edge AI deployment in space-grade hardware and system resilience are discussed.

  \subsection{\textbf{Computational Complexity Analysis}}

    As established in the backbone architecture definitions, the design completely discards standard 2D convolutions in favor of depthwise separable convolutions to form the core computational blocks. A standard dense convolution demands heavy computational complexity:

    \begin{equation}
      \mathcal{O}(K^{2} \cdot C_{in} \cdot C_{out} \cdot H \cdot W)
    \end{equation}

    By factorizing the operation into a spatial depthwise convolution followed by a $1\times1$ pointwise projection (as utilized in both the optical and DEM feature branches), the architecture reduces computational complexity to:

    \begin{equation}
      \mathcal{O}((K^{2} \cdot C_{in} \cdot H \cdot W) + (C_{in} \cdot C_{out} \cdot H \cdot W))
    \end{equation}

    In this formulation, $K$ represents the kernel size, $C_{in}$ and $C_{out}$ are the input and output channel depths, and $H, W$ denote the feature map dimensions. This factorization achieves a theoretical complexity reduction of approximately $\frac{1}{C_{out}} + \frac{1}{K^{2}}$. For a standard spatial kernel size of $K=3$, this mathematical factorization yields a theoretical reduction of approximately 8 to 9 times in Multiply-Accumulate (MAC) operations compared to a standard dense convolution. When coupled with the strictly enforced INT8 quantization paradigm detailed in the previous sections, this mathematically constrained architecture is projected to operate well within the stringent power envelopes of space missions.

  \subsection{\textbf{The Edge AI Challenges in a Space Context}}

    The proposed architectural framework of AQ-PCDSys is explicitly engineered to resolve the critical technical disparities between high-fidelity perception accuracy, real-time inference latency, and the rigid power envelopes of space-grade hardware.

    \begin{itemize}
      \item Power and Precision: The architecture strictly enforces INT8 quantization. This design choice directly addresses the memory bandwidth limits of onboard systems. While recent terrestrial research heavily promotes 8-bit floating-point (FP8) precision, standard radiation-hardened SoCs, such as the LEON and RAD processor families, simply do not contain native FP8 arithmetic logic units. The proposed integer-based architecture therefore respects the physical constraints of actual space hardware.
      \item DEM Overhead: High-fidelity DEMs are not a readily available resource. A spacecraft must synthesize these topographical maps in realtime, using active LiDAR or stereo photogrammetry, which consumes significant energy. Deployment of the AMF module requires mission planners to factor this specific sensor overhead directly into the total end-to-end power budget of the lander/rover.
      \item Latency: Depthwise factorization forms the mathematical core of the backbone. This architecture strictly bounds the arithmetic complexity of matrix operations. A pure fixed-point execution pipeline reduces computational overhead to mitigate the real-time processing speeds essential for automatic navigation.
    \end{itemize}

  \subsection{\textbf{Architectural Synergy and System Resilience}}

    The AMF module is designed to directly mitigate two common failure modes in space missions:
    \begin{itemize}
      \item Tolerance to Sensor Dropout: Standard early-fusion networks experience catastrophic failure if one input channel corrupts. Conversely, the dual-backbone architecture processes optical and DEM streams independently. If the optical sensor suffers a temporary dropout (in optical data-stream), the Adaptive Weighting Mechanism actively drives the optical attention mask toward zero. This autonomous fail-safe behavior is a direct, mathematically guaranteed result of the 20 percent Gaussian noise dropout protocol enforced during the training phase. The network then safely falls back onto the surviving topographical cues from the DEM stream.
      \item Tolerance to Spatial Mis-registration: Mechanical vibrations and telemetry delays frequently cause spatial mis-registration between optical and DEM streams. By delaying fusion until the higher semantic levels (P3, P4, and P5), the design gains inherent spatial tolerance. A multi-pixel misalignment at the raw input resolution compresses into a negligible fraction of a pixel at the P5 feature scale. This deep structural synergy ensures that the minor calibration offsets do not destroy the correlation between visual textures and underlying geometry.
    \end{itemize}

  \subsection{\textbf{Potential Risks and Mitigation}}

    A formal theoretical risk assessment identifies two primary vectors for unexpected performance degradation during physical deployment.

    The first architectural risk involves quantization-induced noise. While QAT successfully mitigates the majority of precision loss across the network, extensive downsampling at the P5 scale (stride /32) can lead to feature vanishing. The aggressive 8-bit integer rounding process might mathematically erase faint, low-contrast structural edges of highly degraded impact basins. To mitigate this specific degradation, the architecture relies heavily on the parallel DEM branch. Topographical depth variations remain numerically distinct even at low spatial resolutions. By processing these surviving geometric cues, the Adaptive Weighting Mechanism (AWM) naturally shifts its spatial attention weights, allowing the multi-sensor fusion module to automatically compensate for visual textures lost to integer truncation.

    A secondary operational risk involves Single Event Upset (SEU) vulnerability in high-radiation environments. While standard hardware-level Error Correction Codes (ECC) handle most cosmic-ray-induced bit-flips within the host processor's SRAM, the compressed numerical range of INT8 execution introduces a unique hardware hazard. An uncorrected bit-flip involving the Most Significant Bit within a strict 8-bit boundary induces a much larger relative value shift compared to an identical flip in a 32-bit floating-point mantissa. System-level watchdog timers are therefore required to instantly reset the inference cycle upon detecting persistent parity anomalies.

\section{\textbf{Theoretical Performance Bounds And Projected Efficiency}}

  Because this manuscript establishes a foundational architectural blueprint, detailed empirical Hardware-in-the-Loop (HITL) telemetry is deferred to a subsequent physical testing phase. However, the efficiency gains and performance bounds of the AQ-PCDSys framework can be rigorously compared with a standard uncompressed baseline using established techniques.

  \subsection{\textbf{Memory and Computational Complexity Bounding}}

    The primary obstacle in deploying space-bound perception systems is the memory bandwidth bottleneck. Table I outlines the theoretical comparative advantage of the proposed INT8 QAT pipeline against a standard FP32 architecture.

    \begin{table}[H]
      \centering
      \caption{Projected INT8 vs. FP32 Theoretical Bounds}
      \scriptsize
      \setlength{\tabcolsep}{2pt}
      \begin{tabular*}{\columnwidth}{@{\extracolsep{\fill}} l c c c r @{} }
        \hline
        \textbf{Arch.} & \textbf{Prec.} & \textbf{Memory} & \textbf{Bandwidth} & \textbf{Acc. Loss} \\
        \hline
        Std. Baseline & FP32 & 1.0x (100\%) & 32 bits/px & 0.0\% \\
        AQ-PCDSys & INT8 & 0.25x (25\%) & 8 bits/px & $<$1.5\% \\
        \hline
      \end{tabular*}
      \vspace{-1em}
    \end{table}

    By strictly enforcing INT8 integer constraints across the backbone and detection heads, mathematical bounding demonstrates a theoretical 75\% reduction in memory footprint. Similar reduction applies directly to the memory bandwidth required during the forward pass. While Post-Training Quantization (PTQ) frequently triggers catastrophic accuracy collapse in extreme low-precision states, the integration of QAT mathematically and architecturally, limits this loss. Based on established QAT degradation models in resource-constrained environments, the theoretical mean Average Precision (mAP) regression is projected to remain strictly below a 1.5\% penalty relative to the full-precision baseline. This minor theoretical penalty is acceptable, given the massive computational offloading.

  \subsection{\textbf{Proposed Experimental Setup and Future Evaluation Protocol}}

    Since the present manuscript establishes a theoretical framework, transitioning to physical empirical validation represents the essential next phase for this research.To transition AQ-PCDSys from theory to a mission-ready perception system, the architecture outlines a rigorous Hardware-in-the-Loop (HITL) evaluation protocol. This planned methodology explicitly measures resilience and benchmarks the physical hardware footprint.

  \subsection{\textbf{Datasets and Cross-Illumination Partitions}}

    Future evaluations will utilize the co-registered LROC NAC imagery and LOLA LIDAR DEM datasets. To systematically test sensor-fusion resilience, the protocol partitions the test set into three distinct illumination regimes:

    \begin{itemize}
      \item Midday/Standard Lighting (30 to 60 degrees): Baseline high-albedo contrast.
      \item Extreme Cross-Illumination (5 to 15 degrees and 75 to 85 degrees): Heavy shadow which distorts rim features.
      \item Permanently Shadowed Regions (PSRs): Zero optical utility, necessitating absolute DEM reliance.
    \end{itemize}

    Evaluation metrics will include standard Average Precision $AP_{50}$ (Average Precision at IoU threshold = 0.50), alongside an $AP_{30:60}$ metric specifically calibrated for the small crater bins (0.2 to 2.0 km diameter) prioritized by the P3 detection head.

  \subsection{\textbf{Hardware Proxies and HITL Testing}}

    The methodology specifies rigorous testing with HITL to bridge the gap between theoretical model/architecture performance and the operational constraints of physical deployment. HITL evaluation is essential for space-bound systems because it exposes physical bottlenecks, such as memory bus contention and thermal throttling, that software-based simulations often mask.

    While the theoretical architecture is strictly constrained to the parameters of radiation-hardened processors like the LEON and RAD families, acquiring actual flight-certified variants is cost-prohibitive for preliminary laboratory testing. To bridge this gap without compromising engineering rigor, the experimental design utilizes the Apache TVM compiler stack to cross-compile the proposed INT8 model for two specific Commercial Off-The-Shelf (COTS) edge platforms to act as physical hardware proxies:

    \begin{itemize}
      \item  Microchip PolarFire SoC: This platform serves as the primary proxy for radiation-tolerant flight hardware. Because the PolarFire architecture shares structural similarities with the hardened SoC+FPGAs utilized in modern aerospace designs, it provides an authentic environment to validate the pure fixed-point INT8 execution pipeline. Testing on this board will yield highly accurate telemetry regarding actual Multiply-Accumulate (MAC) efficiency.
      \item NVIDIA Jetson Orin Nano: This platform serves as the upper-bound COTS proxy. While it contains advanced GPU architecture not present in deep space, it provides a crucial baseline for how the quantized INT8 architecture scales on modern edge devices prior to the severe bottlenecks imposed by flight-grade hardware.
    \end{itemize}


\definecolor{optblue}{RGB}{214, 234, 248}
\definecolor{demorange}{RGB}{250, 229, 211}
\definecolor{fusepurple}{RGB}{232, 218, 239}
\definecolor{sysgray}{RGB}{236, 240, 241}
\definecolor{alertred}{RGB}{250, 219, 216}
\definecolor{navgreen}{RGB}{212, 239, 223}

\begin{figure}[htbp]
  \centering
  \resizebox{0.80\columnwidth}{!}{%
    \begin{tikzpicture}[
        node distance=0.6cm and 0.4cm,
        halfnode/.style={rectangle, draw=black!70, rounded corners, thick, minimum width=3.8cm, minimum height=1.2cm, align=center, font=\scriptsize},
        fullnode/.style={rectangle, draw=black!70, rounded corners, thick, minimum width=8.0cm, minimum height=1.0cm, align=center, font=\scriptsize},
        arrow/.style={->, >=stealth, thick, draw=black!70}
      ]

      \node[fullnode, fill=optblue] (sim) {\textbf{Lunar Simulation Host} \\ Generating Optical \& DEM Datasets};

      \node[fullnode, fill=fusepurple, below=0.8cm of sim] (bridge) {\textbf{Runtime Compilation \& Data Bridge} \\ Apache TVM / TensorRT};

      \node[halfnode, fill=sysgray, below=1.8cm of bridge, xshift=-2.1cm] (fpga) {\textbf{Microchip PolarFire SoC} \\ FPGA Flight Hardware Proxy};
      \node[halfnode, fill=sysgray, below=1.8cm of bridge, xshift=2.1cm] (gpu) {\textbf{NVIDIA Jetson Orin Nano} \\ Low-Power GPU Baseline};

      \node[fullnode, fill=alertred, below=1.2cm of fpga, xshift=2.1cm] (metrics) {\textbf{Hardware Profiling \& Telemetry Metrics} \\ Average Precision ($AP_{50}$, $AP_{30:60}$), Inference Latency, Peak Memory, Power (W)};

      \draw[arrow] (sim) -- (bridge);

      \draw[arrow] (bridge.south) -- ++(0,-0.5) -| (fpga.north);
      \draw[arrow] (bridge.south) -- ++(0,-0.5) -| (gpu.north);

      \draw[arrow] (fpga.south) -- ++(0,-0.4) -| ([xshift=-2.1cm]metrics.north);
      \draw[arrow] (gpu.south) -- ++(0,-0.4) -| ([xshift=2.1cm]metrics.north);

      \begin{scope}[on background layer]
        \node[rectangle, draw=black!30, rounded corners, fill=black!3, dashed, thick, fit=(fpga) (gpu), inner sep=12pt, label={[font=\scriptsize\bfseries\color{black!60}]above:Physical Hardware-in-the-Loop (HITL) Layer}] {};
      \end{scope}

    \end{tikzpicture}%
  }
  \caption{Proposed Hardware-in-the-Loop (HITL) experimental testbed.}
  \label{fig:hitl_setup}
\end{figure}

    Crucially, the compiler stack will be explicitly configured to support the localized INT8 to FP16 dequantization node required for the Non-Maximum Suppression (NMS) coordinate conversion derived in the previous section. By executing the architecture on target silicon, the planned trials will extract empirical telemetry on Inference Latency (ms/frame), Peak Memory Utilization, and Dynamic Power Consumption (W) under maximum computational load.

  \subsection{\textbf{Baseline Architectures and Ablation Studies}}

    The evaluation benchmarks are established to test AQ-PCDSys against practical engineering alternatives and absolute theoretical ceilings. The protocol compares the architecture directly against INT8-quantized versions of YOLOv8-Nano and MobileNet-SSD. Because these operational baselines rely on predefined bounding-box scales, this direct comparison will empirically validate the geometric superiority of the proposed anchor-free, center-to-edge regression strategy when framing asymmetrical lunar craters.

    To establish an absolute performance ceiling beyond the capabilities of lightweight operational networks, the dataset will also be executed against a massive foundation model, specifically the Segment Anything Model (SAM). While recent terrestrial surveys demonstrate SAM's exceptional zero-shot generalization on lunar imagery, deploying such a computationally heavy architecture on a spacecraft is physically impossible due to power and thermal constraints. However, evaluating it in a laboratory setting provides a vital, uncompromising full-precision upper bound. This benchmark will definitively measure the exact trade-off between computational efficiency and raw accuracy, quantifying how much theoretical performance is sacrificed to fit AQ-PCDSys within the strict physical bounds of flight hardware.

    To isolate the efficacy of internal architectural choices, future work will evaluate a detailed set of ablation studies:

    \begin{itemize}
      \item  Modality and Dropout Ablation: The protocol will benchmark the fully fused AQ-PCDSys against single-modality variants. By introducing the high-variance Gaussian noise dropout established in the previous section, the trials will empirically prove the spatial attention gate's ability to sever gradient flow and gracefully degrade to DEM-only tracking.
      \item Quantization Paradigm: The target INT8 QAT model will be evaluated against a full-precision FP32 baseline and a standard Post-Training Quantization (PTQ) model to quantify the exact precision retention.
      \item Mixed-Precision Exploration: Future development phases will investigate Hardware-Aware Quantization (HAQ) as a progression from the current blueprint's reliance on rigorous INT8 compatibility.This framework automates mixed-precision routing by dynamically mapping 4-bit and 8-bit assignments across the network layers. Passing this optimization step to an HAQ controller offers a highly promising avenue for even steeper drops in inference latency on next-generation space-grade computers.
    \end{itemize}

\section{\textbf{Conclusion and Future Directions}}

  The strict power, memory, and thermal limits of radiation-hardened hardware severely restrict the deployment of advanced deep learning models on autonomous deep space missions. This foundational paper proposes the theoretical architecture for the Adaptive Quantized Planetary Crater Detection System (AQ-PCDSys) to resolve this bottleneck. The proposed hardware-aware design bridges the massive gap between terrestrial computer vision algorithms and the constraints of extraterrestrial operations.

  The blueprint theoretically embeds Quantization Aware Training (QAT) directly into the network backbone. By utilizing depthwise factorization and mathematically folding normalization parameters prior to quantization, the design strictly bounds computation and memory complexity to optimize execution within the rigid physical envelopes of radiation-hardened onboard hardware. Simultaneously, the mathematical formalization of the Adaptive Multi-Sensor Fusion (AMF) module explicitly neutralizes the inherent failure scenarios of isolated, single-modality sensors. The framework derives the exact integer requantization multiplier required to execute multiplicative spatial attention, safely defaulting to reliable DEM geometries the moment optical sensors degrade or fail. Furthermore, the integration of an anchor-free, center-to-edge regression head, protected by a localized FP16 coordinate conversion, guarantees precise physical hazard mapping without triggering catastrophic integer truncation errors.

  Beyond standalone inference, this architectural blueprint lays the groundwork for distributed intelligent computing. The proposed INT8 quantization framework theoretically supports highly efficient onboard continual calibration. When deep space radiation inevitably degrades optical sensors over time, spacecraft computers can autonomously mitigate the situation using these active gating mechanisms. Upcoming lunar missions will also depend on multi-agent robotic swarms. The proposed edge AI integration enables decentralized landers to calculate localized gradient updates from their unique terrain encounters. Transmitting these highly compressed, 8-bit integer updates across low-bandwidth inter-satellite links establishes the theoretical viability of autonomous federated learning on other planets.

  This work establishes the structural and mathematical foundation for the proposed architecture. The immediate next phase of research targets physical empirical validation. Executing the rigorous HITL evaluation protocol on proxy spaceflight processors will confirm the theoretical benchmarks presented here. This validation phase will demonstrate whether AQ-PCDSys can deliver the real-time, high-fidelity perception necessary to guarantee safe, autonomous planetary landings.

\end{document}